\documentclass[letterpaper, 10 pt, conference]{ieeeconf}  

\IEEEoverridecommandlockouts                              

\usepackage{booktabs} 
\usepackage{graphicx, caption, subcaption}
\usepackage{flushend}

\usepackage{amsmath}
\usepackage{algorithm}
\usepackage[noend]{algpseudocode}
\usepackage{siunitx}

\usepackage{amsmath}

\title{\LARGE \bf
Casualty Detection from 3D Point Cloud Data \\for Autonomous Ground Mobile Rescue Robots
}

\author{Roni Permana Saputra$^{1,2}$ and Petar Kormushev$^{1}$
\thanks{}
\thanks{$^{1}$Roni P. Saputra and Petar Kormushev are with the Robot Intelligence Lab, Dyson School of Design Engineering, Imperial College London, 
        UK
        {\tt\small \{r.saputra, p.kormushev\}@imperial.ac.uk}}%
\thanks{$^{2}$Roni P. Saputra is also with the Research Center for Electrical Power and Mechatronics, Indonesian Institute of Sciences - LIPI,
        Indonesia}%
}

\begin{document}

\maketitle
\thispagestyle{empty}
\pagestyle{empty}

\begin{abstract}

One of the most important features of mobile rescue robots is the ability to autonomously detect casualties, i.e. human bodies, which are usually lying on the ground. This paper proposes a novel method for autonomously detecting casualties lying on the ground using obtained 3D point-cloud data from an on-board sensor,  such as an RGB-D camera or a 3D LIDAR, on a mobile rescue robot. In this method, the obtained 3D point-cloud data is projected onto the detected ground plane, i.e. floor, within the point cloud. Then, this projected point cloud is converted into a grid-map that is used afterwards as an input for the algorithm to detect human body shapes. The proposed method is evaluated by performing detections of a human dummy, placed in different random positions and orientations, using an on-board RGB-D camera on a mobile rescue robot called ResQbot. To evaluate the robustness of the casualty detection method to different camera angles, the orientation of the camera is set to different angles. The experimental results show that using the point-cloud data from the on-board RGB-D camera, the proposed method successfully detects the casualty in all tested body positions and orientations relative to the on-board camera, as well as in all tested camera angles.

\end{abstract}

\section{INTRODUCTION}
Searching and rescuing injured humans, i.e. casualties, in a disaster situation is one of the key challenges for autonomous mobile rescue robots in search and rescue (SAR) missions. In order to operate autonomously in a SAR mission, a mobile rescue robot needs to be able to detect human casualties unassisted. In these disaster scenarios, the casualty detection problem causes a wide range of challenges, including environmental challenges and a huge amount of different human body poses, compared to the state-of-the-art human-presence detection methods.

A number of successful research studies have been conducted for human-presence detection methods [1][2]. Most of these studies rely on 2D image information and classification to categorise and localise human presence in the image scene. These methods have been reported as successfully dealing with object categorisation and detecting human presence in many different cases.

Using 3D shape information, such as point-cloud data, could also be an alternative for dealing with the human-presence detection problem. This point-cloud data can be obtained from several different sensors, such as RGB-D cameras and 3D LIDAR sensors. A number of studies have looked into detecting human presence from 3D point-cloud data and have reported successful implementations [3][4].

\begin{figure}[t]
\centering
\includegraphics[width=3.4in]{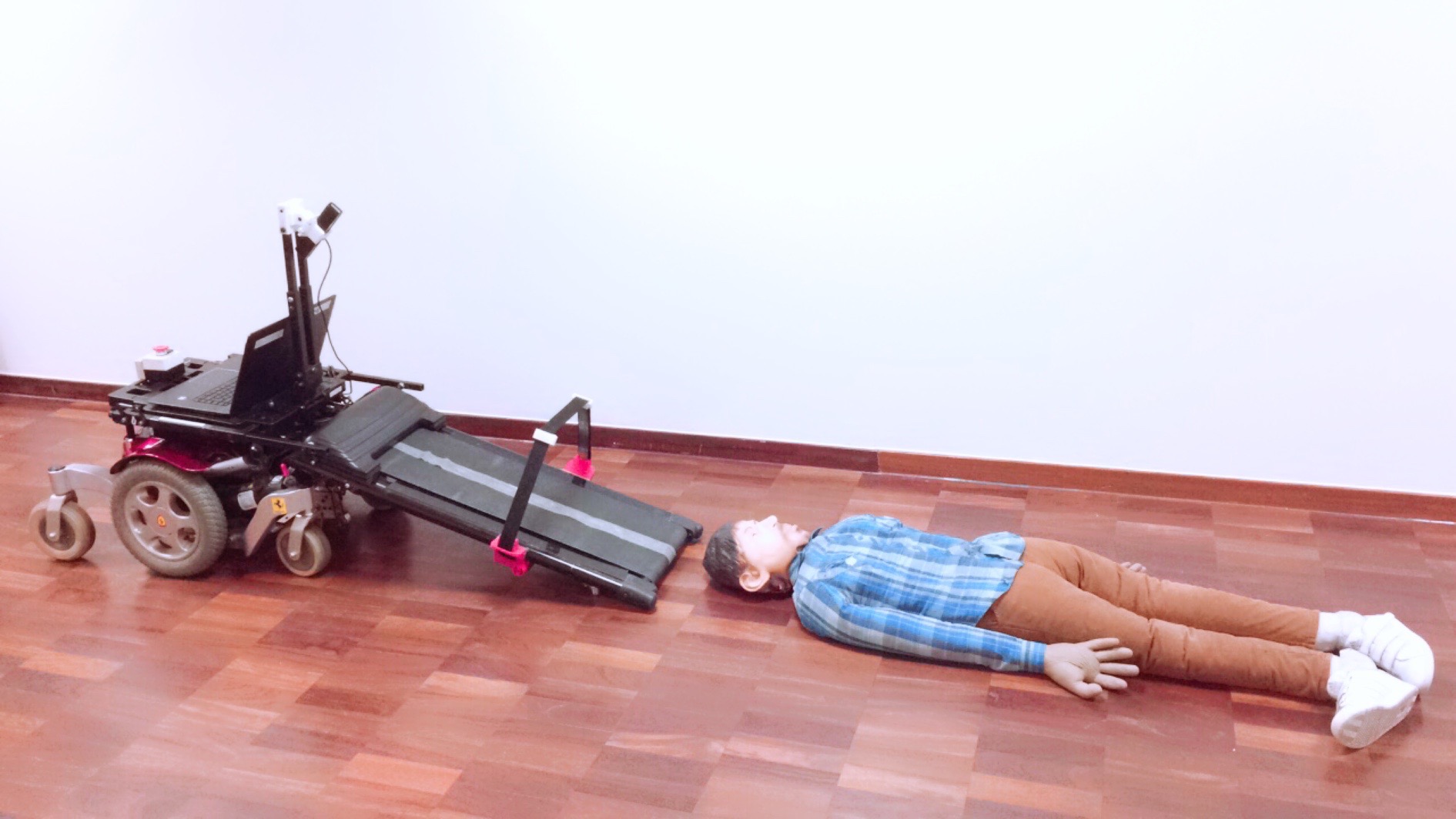}
\caption{Autonomous mobile rescue robot ResQbot [21] for casualty extraction scenarios. This platform comprises a motorised stretcher bed conveyor module attached to a differential-drive mobile base, a stretcher strap module and an RGB-D camera.}
\label{fig:resqbot}
\end{figure}

Despite the number of successful state-of-the-art implementations, using the existing methods for detecting casualties, i.e. human bodies in disaster scenes, remains challenging. The reasons for this are:
\begin{enumerate}
\item It is very difficult to recognise and detect the human body in all situations using 2D image inputs. This is due to the huge amount of variation caused by lighting, clothing, hair style, camera angles, and image resolutions;
\item Most of the reported approaches in the literature either deal with detecting certain parts of the human body or limiting the search to specific poses only, e.g. standing positions;
\item The fact that the casualty is lying down on the ground leads to diverse orientations compared to a standing person with respect to the camera; 
\item In terms of methods based on the use of depth information to detect human presence, the fact that the floor is right under the body makes depth segregation extremely challenging; 
\item The viewing angle camera and the robot orientation in 3D images are arbitrary as they frequently change depending on the terrain.   
\end{enumerate}

In this study, we propose a novel approach for detecting a casualty, i.e. a human body in a disaster scene that is lying on the ground, using 3D point-cloud input data. In this approach, the 3D point-cloud data is projected onto the detected ground plane, i.e. floor, within the point clouds. These ground-projected point clouds are then converted into a grid map that can be used as input for the algorithm for detecting human body shapes. 
The contributions of this work include:
\begin{enumerate}
\item A novel approach for detecting a casualty, i.e. a human body in a disaster scene lying on the ground, using 3D point-cloud input data.
\item The evaluation of the proposed approach in relation to detecting a human body lying on the ground in several different random positions and orientations with regards to the RGB-D camera and with several different camera angle settings.
\end{enumerate}

\section{Related Work}
A large number of successful studies have been conducted and reported in the area of human-presence detection methods, specifically methods based on visual information, such as images or videos [5][2]. For many years, the task of detection and recognition of human appearance in image scenes has been of interest to the computer vision research field. A number of research papers have presented various implementations of these human-presence detection techniques in different fields, such as pedestrian detection for autonomous vehicles [6][7], human-motion tracking for surveillance [8], and human-body detection for search and rescue scenarios [9][10][11].

Different techniques have been addressed for dealing with human-presence detection. A very common technique is to use a combination of different feature descriptors and classifiers to detect human presence in images or videos. Some proposed techniques rely on detecting distinct parts of the human body, such as the face, and the other detection techniques specifically train classifiers to detect full human bodies. A very popular technique in computer vision research is the learning-based technique for face detection proposed by Viola and Jones [12]. This method uses an ‘integral image’ as an image representation and a classifier built using the AdaBoost learning algorithm. It has been reported that this method can perform detection for high-frequency image input.

Another popular technique reported in computer science detects full human bodies using grids of histograms of oriented gradient (HOG) descriptors and a linear support vector machine (SVM) classifier as proposed by Dalal and Triggs [13]. According to their experiments, they showed that the HOG descriptor significantly outperforms other state-of-the-art methods that use other feature sets regarding human detection. This technique has been successfully implemented for pedestrian detection and has near-perfect performance in detecting pedestrians from the MIT pedestrian database.

Recent developments in advanced machine learning techniques have led researchers to investigate and propose novel methods for human body detection using deep-learning techniques. In these methods, human-body detection relies on the Convolutional Neural Network (ConvNet) pre-trained with a massive number of labelled images from existing databases, such as the ImageNet database.

Oliveira, et al. propose an approach based on ConvNets for human-body part discovery in conventional RGB images [14]. In this work, a deep architecture of ConvNet was proposed and was trained end-to-end to assign each pixel of the train image to one of a predefined set of human body part classes, such as head, torso, arms, and legs. This method was evaluated using the PASCAL Parts dataset and it demonstrated state-of-the-art performance in human body part discovery from RGB images.

Due to the development of different sensors that provide 3D information, such as 3D LIDAR, stereo vision cameras, and RGB-D cameras, like Kinect, many studies have also looked at human-presence detection using 3D information, such as point-cloud data [15--20]. One benefit of using this 3D information is that it is easier to separate the background information from the region of interest as there is a distinct difference in depth information.

One example of using 3D information for human-body detection was presented by Szarvas, et al. [15]. They demonstrated the use of light detection and ranging (LIDAR) data and ConvNet-based image classifiers for real-time pedestrian detection. They used the range data from the LIDAR sensor to constrain the search region in the image to improve the processing speed of their image classifier. The LIDAR data was used to obtain the region of interest (ROI) on the image and the ConvNet classifier confirmed the presence of the human body in this ROI. These results show that this approach is capable of processing image input in high frequency for detecting pedestrian from these images. 

Another study looking at human-presence detection based on 3D point data was presented by Navarro-Serment, et al. [16]. They utilised 3D measurements from a lased detection and ranging (LADAR) sensor for real-time detection and tracking of pedestrians. They implemented their work for detecting potential pedestrians from an autonomous moving vehicle and used statistical pattern recognition techniques to recognise and classify objects from a subset of point-cloud data obtained from the LADAR sensor mounted on the vehicle. To recognise a human within the objects, the algorithm used geometric and motion features.

Kidono, et al. proposed a novel approach to improve the classification performance of pedestrian detection using high-definition LIDAR data [17]. In their method, they utilised two novel features: the slice feature and the reflection intensities distribution of points measured on the target. To classify these features and detect pedestrian presence, they used the SVM classifier. The evaluation of this method showed that it successfully detected pedestrians in a road environment.

Despite these successful state-of-the-art implementations of human-presence detection, using the existing methods for detecting casualties, i.e. human bodies in disaster scenes, remains challenging. Many issues have to be dealt with in these scenarios, such as the unique environment variation in extreme disaster scenes, the similarity of background scenes and human scenes, and human appearance variability and unpredictability in disaster scenes compared to normal scenes. Creating a classifier for human-body detection that can deal with all these scenarios is very challenging.

For learning-based classifiers, it is possible to customise or extend the classifier by training it with a huge amount of data of the human body in disaster scenarios. This dataset could include human bodies lying on the ground with a huge number of degrees of freedom of the human body pose and the camera viewpoint. To the best of our knowledge, however, there is currently not enough data of the human body in disaster scenarios available. In addition, collecting a huge amount of data for such scenarios is not trivial.

Similar to state-of-the-art human-presence detection based on 2D images, the current existing methods for human detection from 3D information are mostly designed to deal with standard person detection, i.e. detecting people standing or in a normal pose. It is also not trivial to implement the existing techniques for casualty detection in disaster scenarios due to the different characteristics and possibilities of human body poses and positions compared to the common standard person detection cases.

\section{Robot Platform and Hardware}

A mobile rescue robot, called ResQbot (as shown in Figure 1), is used as a robot platform for the experiments in this study. This platform consists of two main modules: a differential-drive mobile base and a motorised stretcher bed conveyor module. This platform was designed and built to be able to perform a safe casualty extraction procedure. In order to perform this casualty extraction routine, this robot adopts a loco-manipulation approach that uses its locomotion system to perform a manipulation task, e.g. in this case, loading the victim from the floor [21][22].

This platform is also equipped with a range of perception devices, including an RGB-D camera and a stereoscopic camera rig. These devices provide the perception required during the operation. For the experiment in this study, we used the ASUS Xtion RGB-D camera on the ResQbot to obtain an RGB image, a depth image, and point-cloud data that could be used for the casualty detection process [23]. The key specification of this RGB-D camera related to the obtained image data can be seen in Table~\ref{tab:rgbd-spec}.

\section{Proposed Approach}

\begin{table}
\begin{center}
  \caption{ASUS Xtion RGB-D camera specifications}
  \label{tab:rgbd-spec}
  \begin{tabular}{ccl}
    \toprule
    Parameters&Value\\
    \midrule
    Distance of Use&between $0.8m$ and $3.5m$\\
    Field of View&$\ang{58}$ H, $\ang{45}$ V, $\ang{70}$ D \\
    &(Horizontal, Vertical, Diagonal)\\
    Depth Image Size&VGA ($640x480$) : $30$fps\\
    &QVGA ($320x240$): $60$fps\\
  \bottomrule
\end{tabular}
\end{center}
\end{table}

\begin{figure}[t!]
\centering
\includegraphics[width=3.4in]{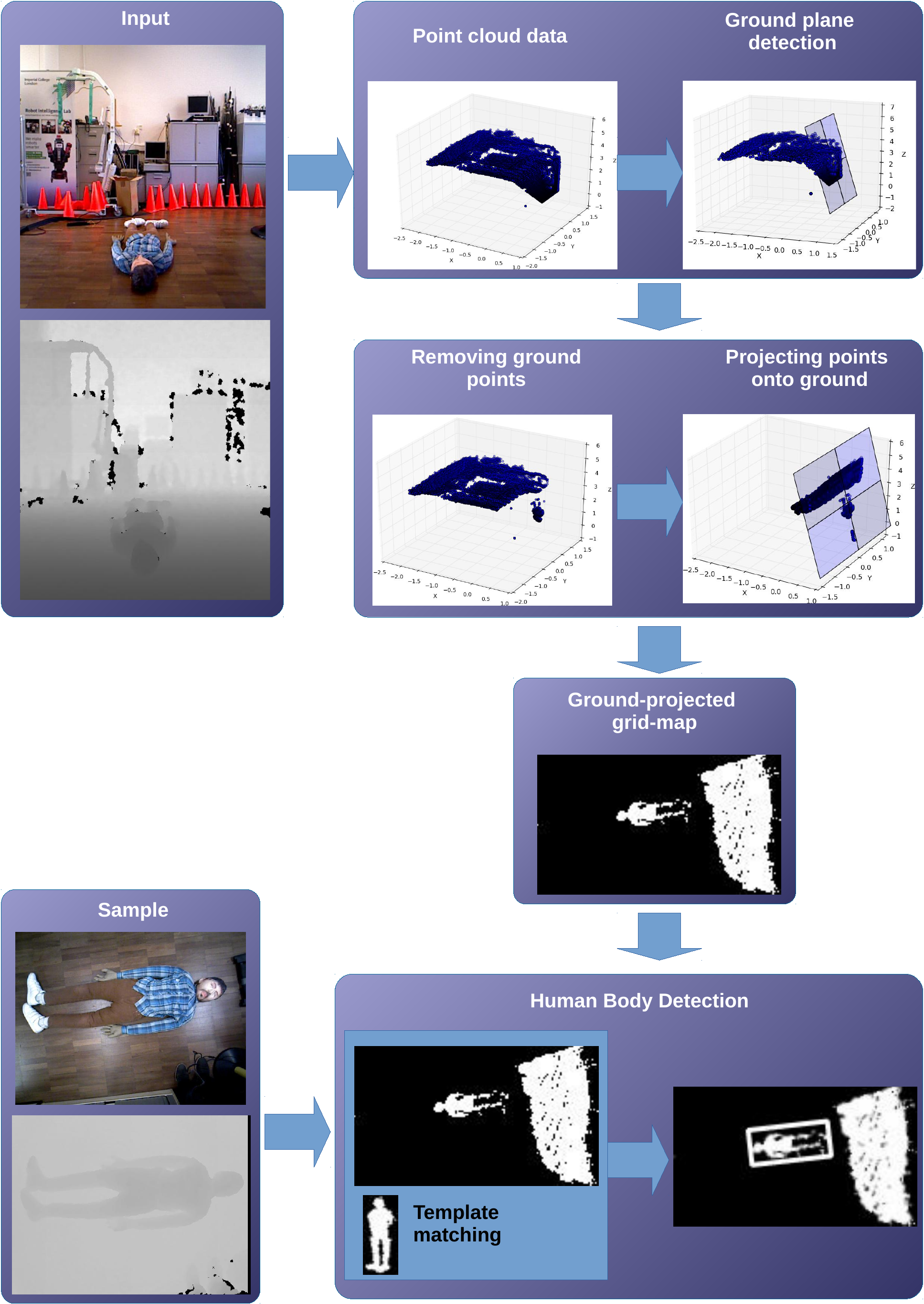}
\caption{Proposed methodology for casualty detection using 3D point-cloud data.}
\label{fig:method}
\end{figure}

\begin{algorithm}[!hb]
\caption{Casualty Detection from Ground Projected Point Cloud}\label{alg:casualty-detect}
\begin{algorithmic}[1]
\Function{DetectCasualty}{}\\
\quad points $\gets$ PointCloudData;\\
\quad Estimate a ground plane using the RANSAC algorithm;\\
\quad Remove all points belonging to the estimated plane;\\
\quad Projecting the remaining points onto the estimated ground plane;\\
\quad Generate a grid map on the basis of the projected points;\\
\quad Perform the object detection procedure to detect a casualty on the grid map;
\EndFunction
\State \Return detected\_casualty
\end{algorithmic}
\end{algorithm}

Figure~\ref{fig:method} illustrates the proposed approach used for casualty detection in this study. In this work, the point-cloud data was obtained from an onboard RGB-D camera mounted on the mobile rescue robot ResQbot. As can be seen in Figure~\ref{fig:method}, we first estimated a ground plane model from the point-cloud data using a random sample consensus (RANSAC) algorithm. Once the ground plane model was detected, all the points belonging to the detected plane were removed.

We then projected all the remaining points onto the detected ground plane model by finding the normal vector towards the plane and along each point. We then generated a grid map with respect to the ground plane using the projected points information. Afterwards, we used the generated grid map as an input for object detection procedure to detect the casualty in the grid map.

Algorithm~\ref{alg:casualty-detect} illustrates the summary of our proposed approach.
In the following subsection, all these steps will be described in more detail.

\subsection{Ground Plane Detection via RANSAC}
The first step in the proposed approach was estimating a ground plane from the point-cloud data input. To estimate this plane, we used the RANSAC algorithm [24][25]. This algorithm is a general model estimation technique designed to deal with an input data with large proportion of outliers. This technique uses the minimum data points to estimates the model parameters, so the process is faster compared to the other common robust estimation techniques [26]. Algorithm~\ref{alg:RANSAC} illustrates the details of the pseudo-code of the RANSAC algorithm.

In general, this algorithm estimates the best model of a plane that fits the point-cloud data input in  $N$ iterations. A random plane model (i.e. plane parameters) is calculated by randomly choosing three points belonging to the point-cloud data. The inlier (i.e. all points from the point cloud belonging to the estimated plane) is then classified by calculating the distance of each point to the plane and comparing the distance to a given threshold. A point belongs to the inlier of a corresponding plane if the distance between this point to the plane is below the given threshold $(t)$. 
These procedures are then repeated for $N$ iterations and the number of inlier points are compared on each iteration. The best fit of the plane model is chosen from the largest number of inlier points from all iterations. The iterations can be stopped if the number of inlier points is large enough or equal to the given threshold.

To ensure the accuracy of the estimation, a large value of $N$ is chosen to ensure that the probability of at least one of the set points, i.e. the three points samples, produced by the random sampling does not include an outlier. Let $N$ represent the number of iterations and m is the minimum number of points required for estimating the model (i.e. three points). If $u$ represents the probability that any selected data point is an inlier, so $(1-u)$ represents the probability of selecting an outlier. The probability that during the sampling process at least one of the sample sets does not include an outlier can be represented as:

\begin{equation}
1-p=(1-u^m)^N
\end{equation}

Thus, the number of required iterations can be approximated by:
\begin{equation}
N=\dfrac{log(1-p)~}{log(1-(1-v)^m)}
\end{equation}

in which the probability $p$ is usually set to $0.99$.

\begin{algorithm}
\caption{Ground Plane Detection via RANSAC}\label{alg:RANSAC}
\begin{algorithmic}[1]
\Function{Estimate Ground Plane}{}
\State best\_inlier $= 0$;
\State num\_iteration $= N$;
\State points $\gets$ PointCloudData;
\State num\_points $= np$;
\State $t =$ maximum inlier distance from plane;
\For{$i$~from~$1$~to~$N$}
\State $p =$ random sampling 3 points
\State PlaneModel $= pm(i) $
\State $pm(i) = $ estimate model from 3 points$(p)$
\State n\_inliers $=0$
\For {$j$ from $1$ to $np$}
\State $d(j) =$ distance point$(j)$ to $pm(i)$
\If {$dis(j)<t$} 
\State point$(j)=$inliers
\State n\_inliers$=$n\_inliers$+1$
\EndIf
\EndFor
\If {n\_inliers$>$best\_inlier}
\State best\_inlier$=$n\_inliers
\State bestPlaneModel$=pm(i)$
\EndIf
\If {best\_inlier \textbf{is} large enough}
\State break
\EndIf
\EndFor
\EndFunction
\State \Return bestPlaneModel, best\_inlier
\end{algorithmic}
\end{algorithm}

\subsection{Point-clouds Orthogonal Projection}
The next step in our approach is projecting the outlier points onto the estimated ground plane. This problem basically is finding the nearest 3D point on the plane to the point of interest. 
Let $X^\prime (x_0, y_0, z_0)$ represent a point within the point clouds and $X (x, y, z)$ represent the nearest point on the plane to point $X^\prime$. Given the estimated ground plane model as:
\begin{equation}
a(x-d)+b(y-e)+c(z-f)=0
\end{equation}
the normal vector to the plane is given by:
\begin{equation}
\textbf{v}=\begin{bmatrix}a\\  b\\ c\\ \end{bmatrix}
\end{equation}

The orthogonal vector from the plane to the point is given by:
\begin{equation}
\textbf{w}=\begin{bmatrix}x-x_0\\  y-y_0\\ z-z_0\\ \end{bmatrix}
\end{equation}
This orthogonal vector $w$ can be also represented in relation to the vector $v$ by:
\begin{equation}
\textbf{w}=\textbf{v}t
\end{equation}
Thus, these vectors can be combined as:
\begin{equation}
\textbf{w}=\begin{bmatrix}at\\  bt\\ ct\\ \end{bmatrix}=\begin{bmatrix}x-x_0\\  y-y_0\\ z-z_0\\\end{bmatrix}
\end{equation}
The parameter $t$ can be calculated as:
\begin{equation}
t=\dfrac{ad-ax+be-by+cf-cz}{a^2+b^2+c^2}
\end{equation}

By substituting the t value into plane equation (7), then we can find the values of $X (x, y, z)$, the orthogonal projection of point $X^\prime (x_0, y_0, z_0)$ on the plane.

\subsection{Grid Map Generation}
\begin{figure}[t]
\centering
\includegraphics[width=2in]{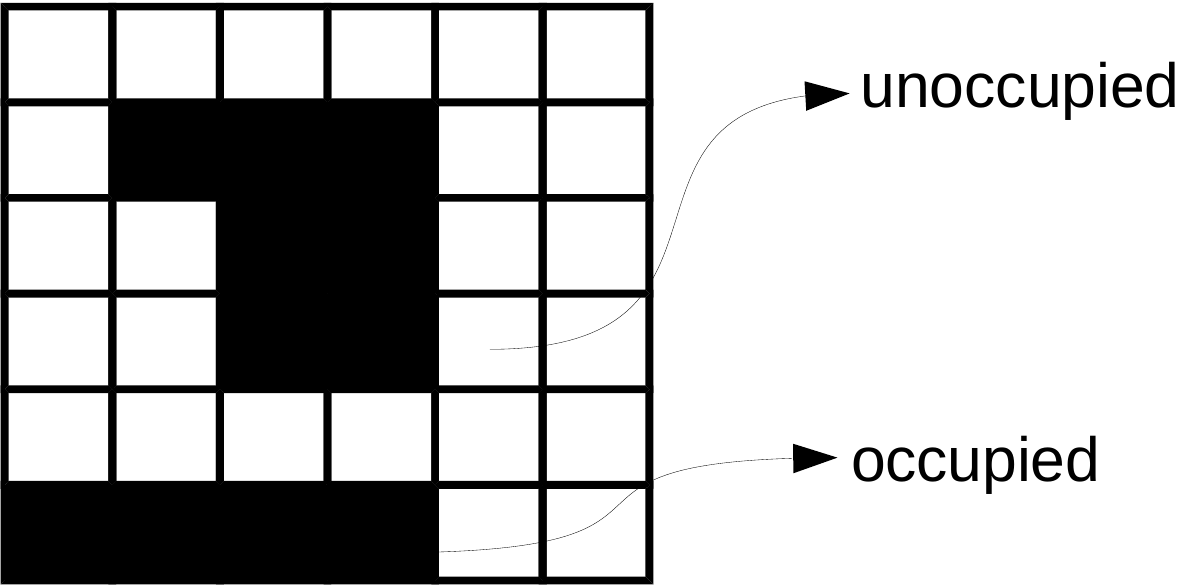}
\caption{Illustration of grid map representation.}
\label{fig:grid}
\end{figure}

After all the points are projected onto the estimated ground plane, a grid map on the basis of these projected points is generated. This grid map is a discrete representation of the workspace. Figure~\ref{fig:grid} illustrates a sample grid map representation. In this work, we generate a binary grid map to represent the projected point. Each cell within the grid is a representation of a binary value $c(i)$. A cell is occupied if at least one projected point is located in the cell.

The value of each grid cell is given by:

\begin{equation}
  c(i)=\begin{cases}
    1, & \text{if cell is occupied}.\\
    0, & \text{otherwise}.
  \end{cases}
\end{equation}

\begin{figure}[b]
\begin{center}
    \begin{subfigure}[c]{0.185\textwidth}
        \includegraphics[width=1.5in]{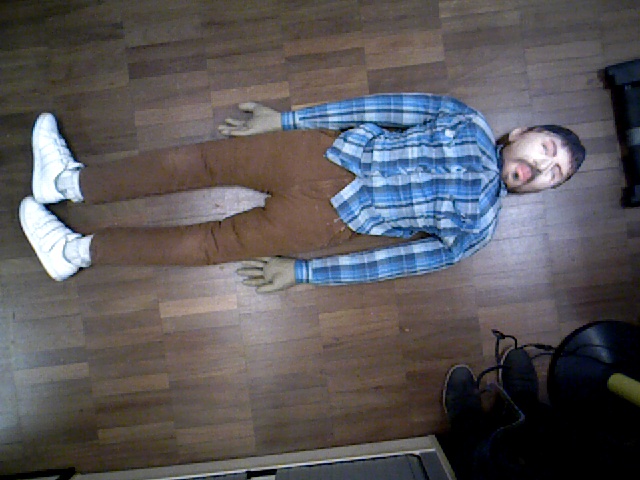} 
        \subcaption{RGB image template}
    \end{subfigure}
    ~~~~
    \begin{subfigure}[c]{0.185\textwidth}
        \includegraphics[width=1.5in]{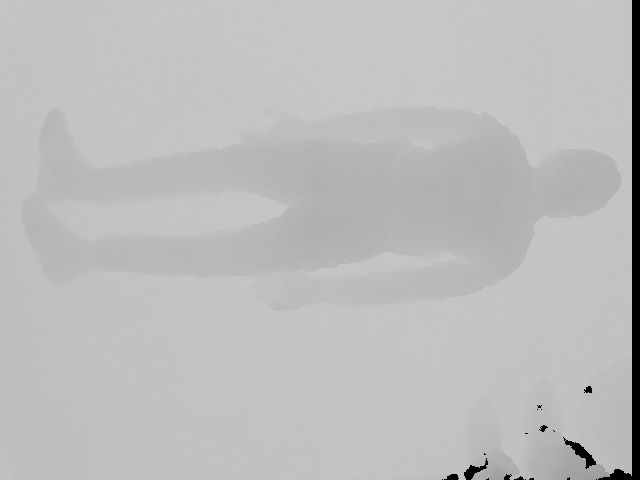}
        \subcaption{Depth image template}
    \end{subfigure}
        ~~~~
    \begin{subfigure}[c]{0.185\textwidth}
        \hspace{1.5pt}
        \includegraphics[width=1.5in]{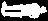}
        \hspace{2pt}
        \subcaption{Grid map template}
    \end{subfigure}
    \caption{Image samples for casualty detection based on template matching.}
    \label{fig:template}
\end{center}
\end{figure} 

\subsection{Human Body Detection from Grid-map}
In this work as a proof-of-concept of our proposed approach, for detecting the human body shape from grid map, we used a standard template matching from OpenCV library [27]. In general, the main goal of the template matching method is to find the highest matching area within the input image, compared to the template image. To identify the matching area, we slide and rotate the template within the image input region, so that the matching process could cope with different orientations as well.

Figure~\ref{fig:template} shows the templates, including an RGB image, a depth image and a generated corresponding grid map, that are used in our experiments. We used the grid map as an input image for template matching process for detecting the human body in the grid map. We only use single image template for the matching process for the whole testing scenarios in this work.

\begin{figure}[t]
\begin{centering}
    \begin{subfigure}[l]{0.078\textwidth}
        \includegraphics[width=.8in]{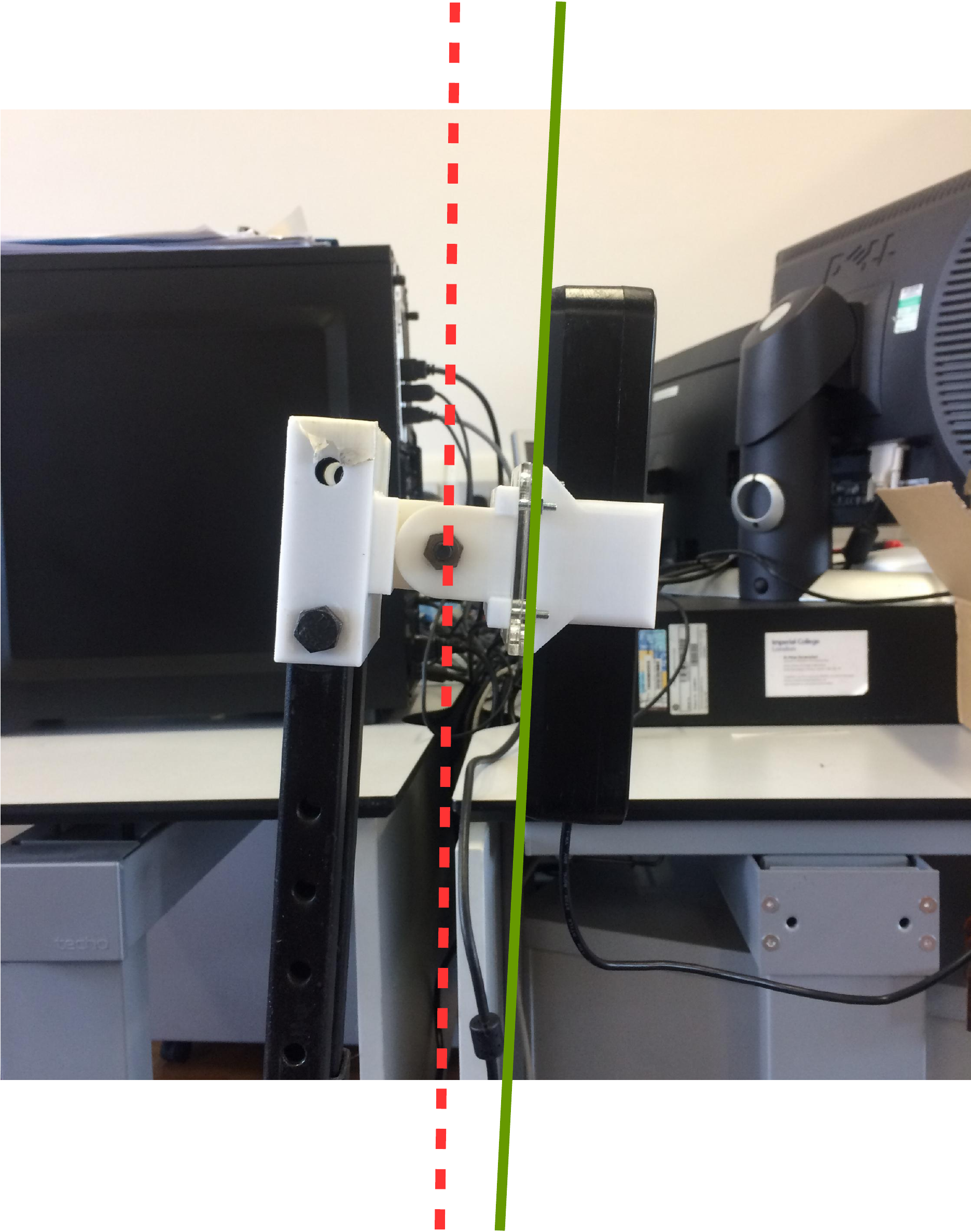} 
        \subcaption{}
    \end{subfigure}
    ~~~~
    \begin{subfigure}[l]{0.078\textwidth}
        \includegraphics[width=.8in]{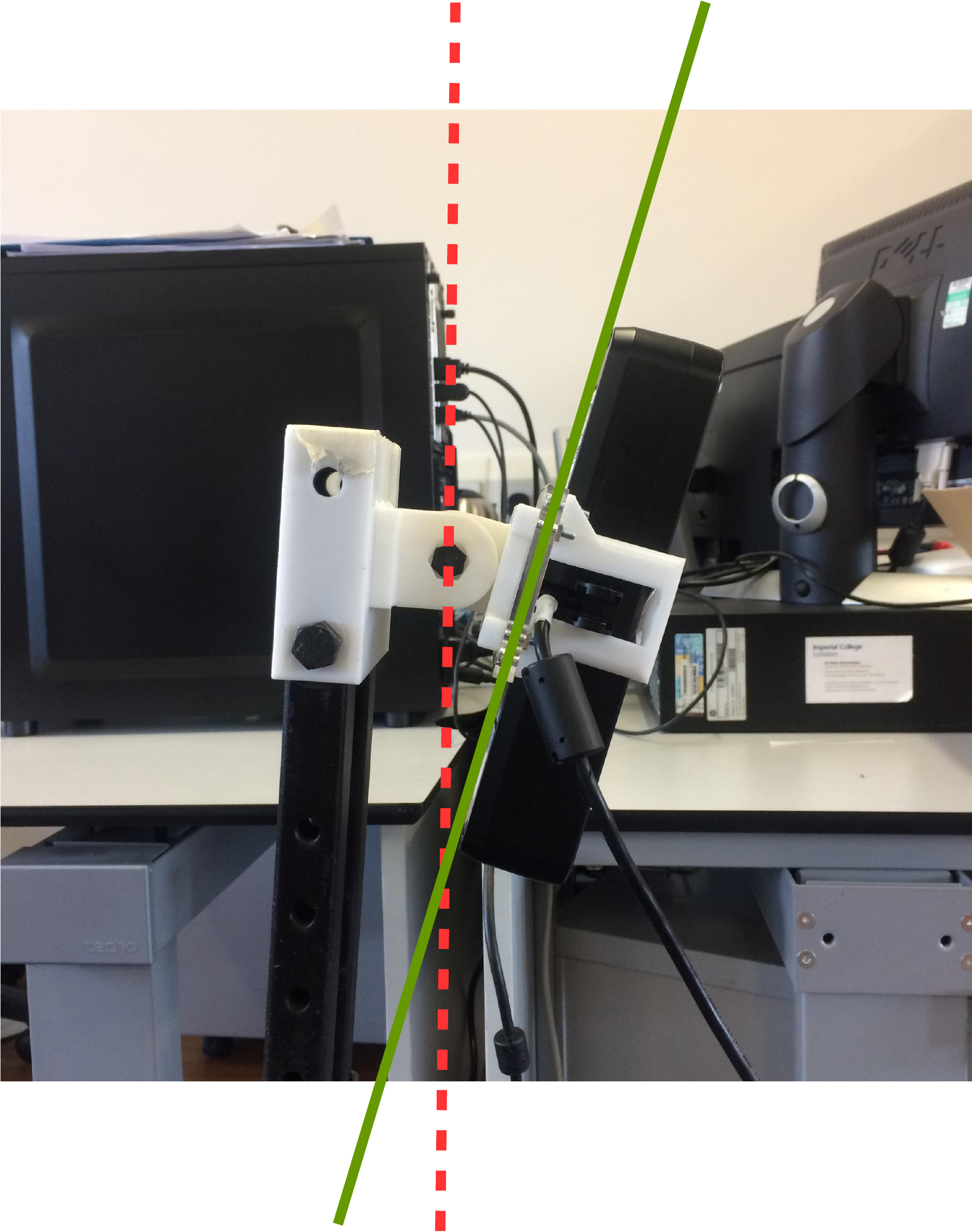}
        \subcaption{}
    \end{subfigure}
    ~~~~
    \begin{subfigure}[l]{0.078\textwidth}
        \includegraphics[width=.8in]{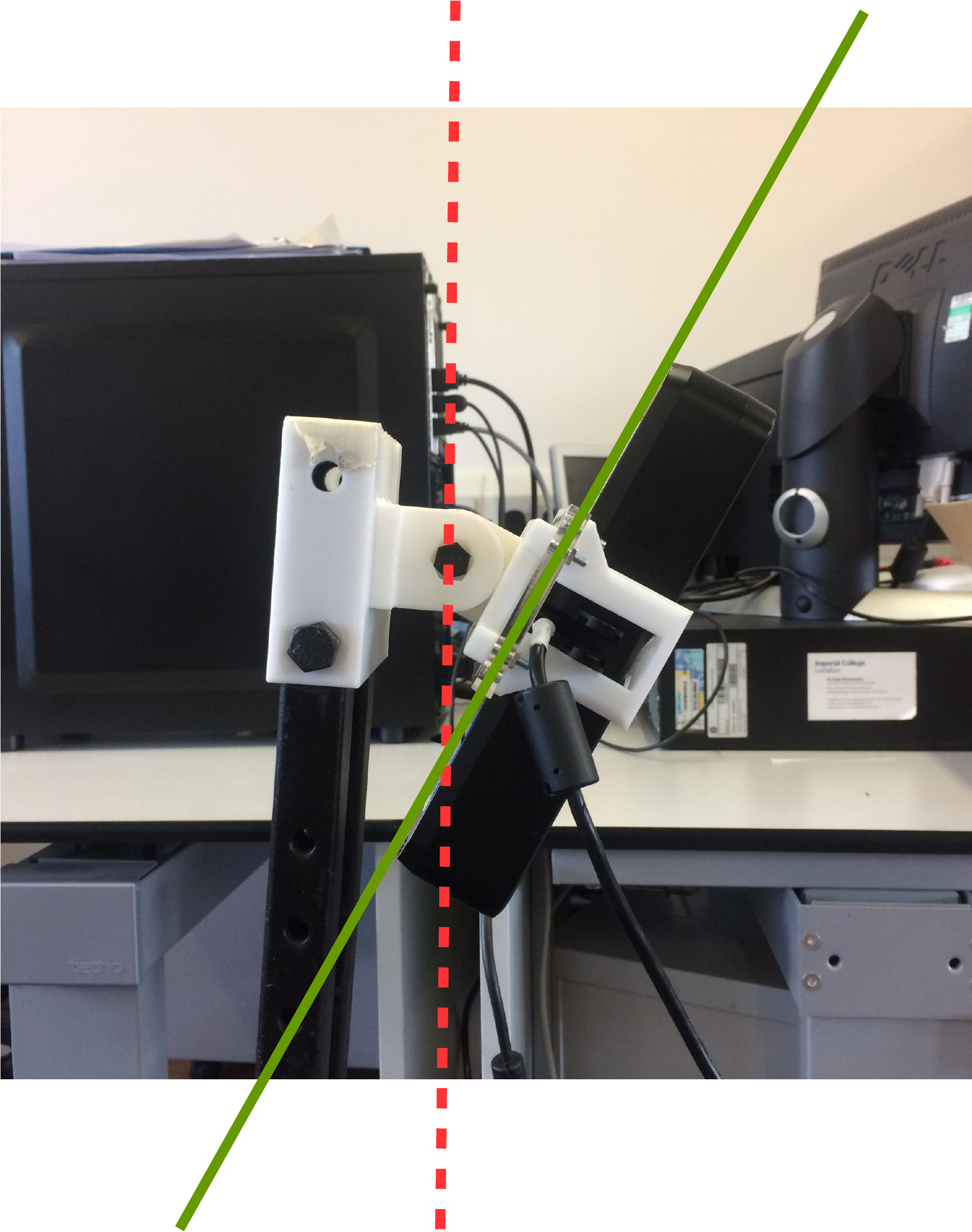}
        \subcaption{}
    \end{subfigure}
    ~~~~
    \begin{subfigure}[l]{0.12\textwidth}
        \includegraphics[width=.8in]{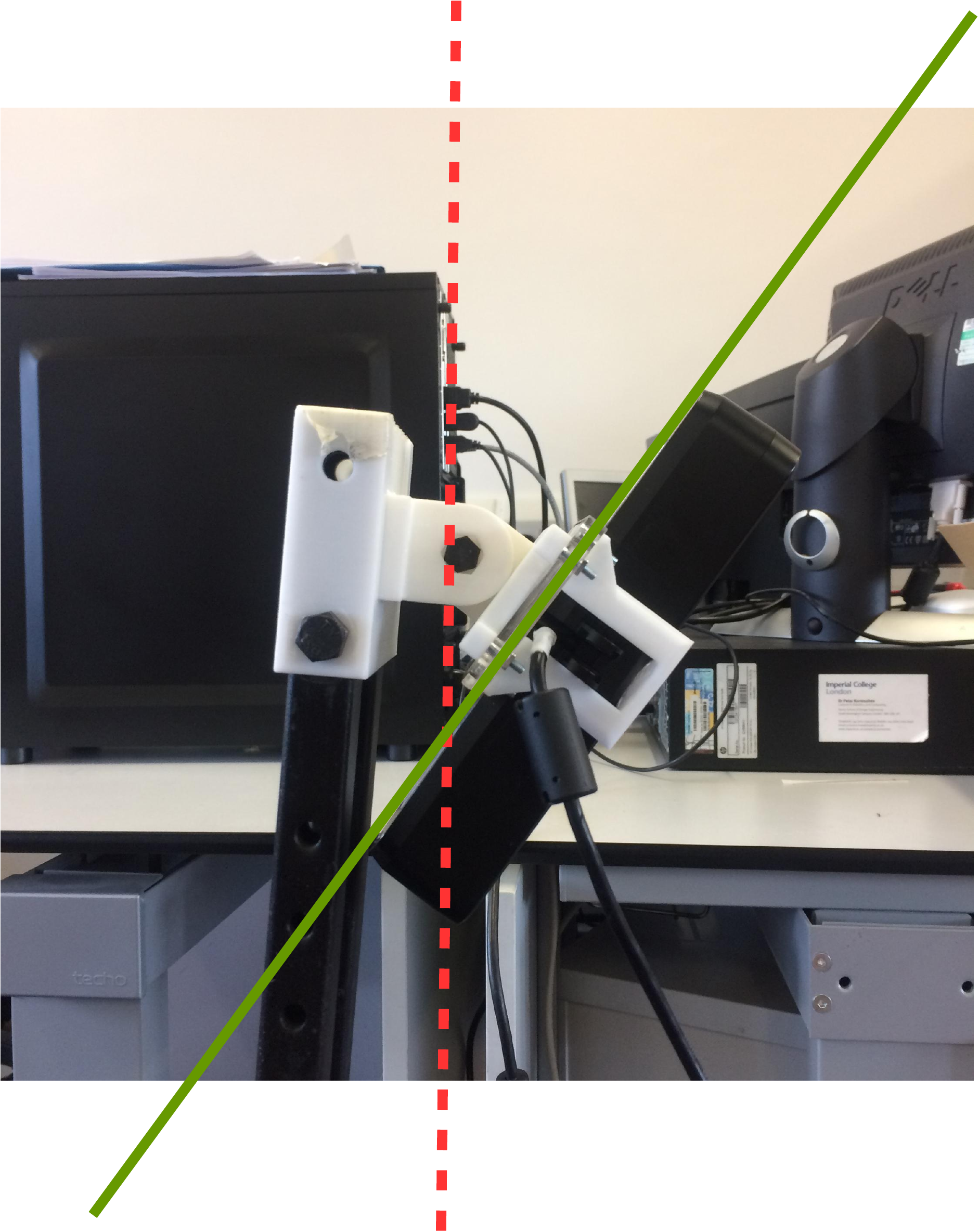}
        \subcaption{}
    \end{subfigure}
    \caption{Four different camera angles.
    Dashed red lines represent the neutral angle (i.e. vertical line) and the continuous green lines represent the camera angle of the RGB-D camera.}
    \label{fig:cam-angle}
\end{centering}
\end{figure}

\section{Experimental Setup and Results}

A set of experiments was conducted to evaluate the proposed casualty detection approach. This approach was tested in a variety of scenarios, including several different human dummy positions and orientations and several different RGB-D camera angle settings. In this experiment, we used four different camera angles (see Figure~\ref{fig:cam-angle}). 

\begin{figure}[t!]
    \begin{subfigure}[c]{0.1\textwidth}
        \includegraphics[width=0.7in]{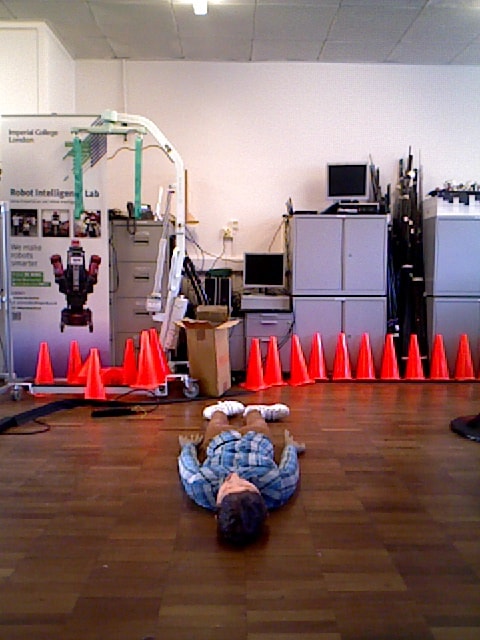} 
        \subcaption{}
    \end{subfigure}
    ~~~~
    \begin{subfigure}[c]{0.1\textwidth}
        \includegraphics[width=0.7in]{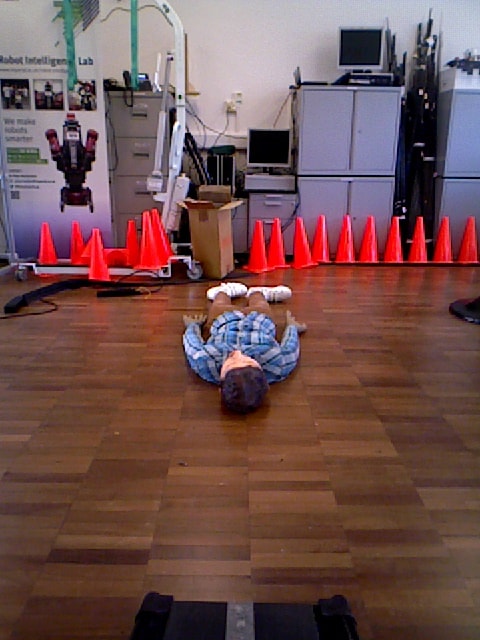}
        \subcaption{}
    \end{subfigure}
    ~~~~
    \begin{subfigure}[c]{0.1\textwidth}
        \includegraphics[width=0.7in]{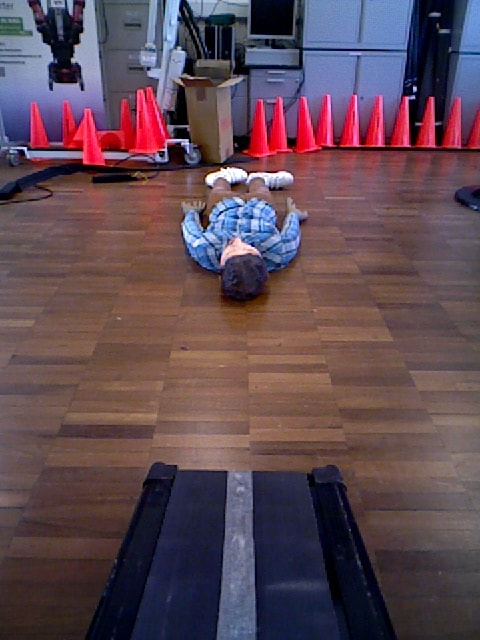}
        \subcaption{}
    \end{subfigure}
    ~~~~
    \begin{subfigure}[c]{0.1\textwidth}
        \includegraphics[width=0.7in]{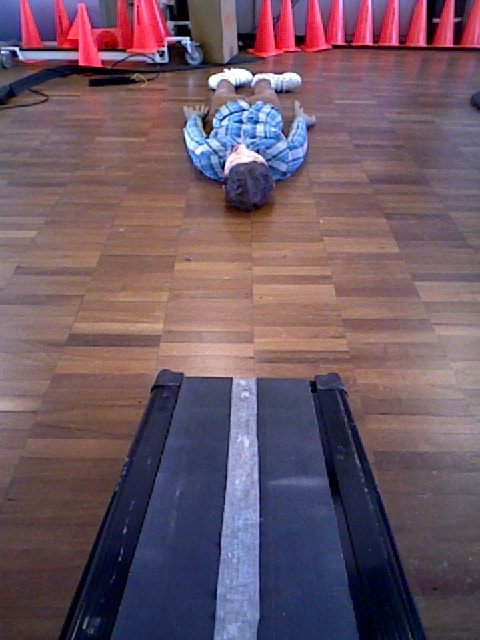}
        \subcaption{}
    \end{subfigure}
    ~~~~
    \begin{subfigure}[c]{0.1\textwidth}
        \includegraphics[width=0.7in]{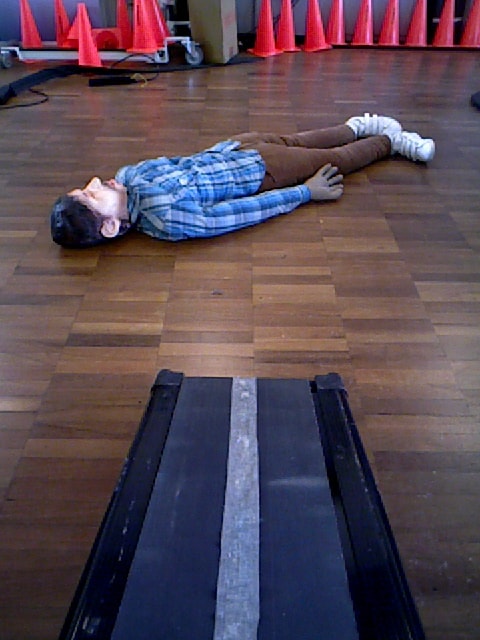} 
        \subcaption{}
    \end{subfigure}
    ~~~~
    \begin{subfigure}[c]{0.1\textwidth}
        \includegraphics[width=0.7in]{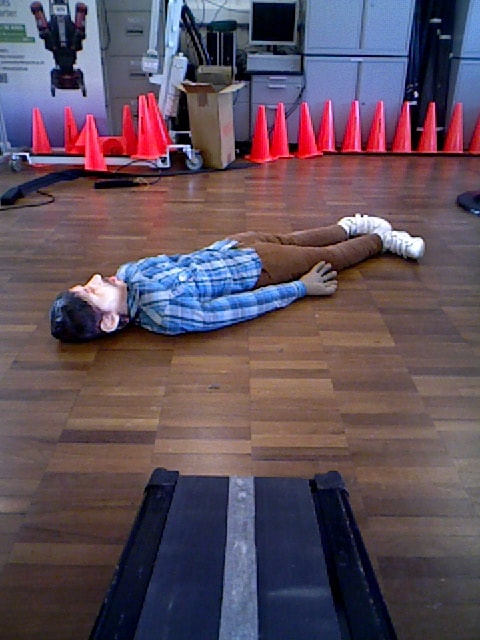}
        \subcaption{}
    \end{subfigure}
    ~~~~
    \begin{subfigure}[c]{0.1\textwidth}
        \includegraphics[width=0.7in]{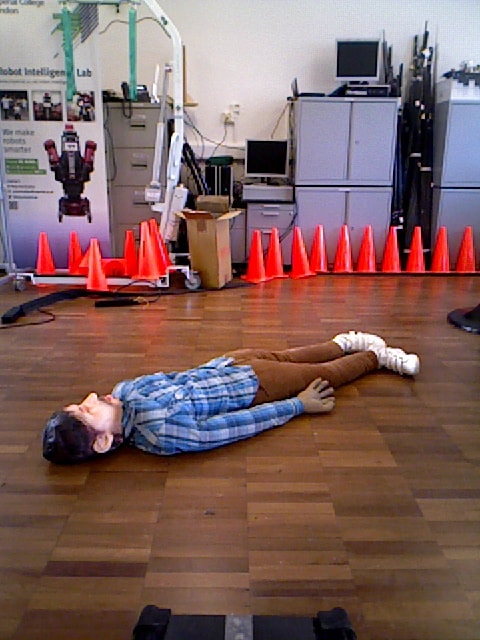}
        \subcaption{}
    \end{subfigure}
    ~~~~
    \begin{subfigure}[c]{0.1\textwidth}
        \includegraphics[width=0.7in]{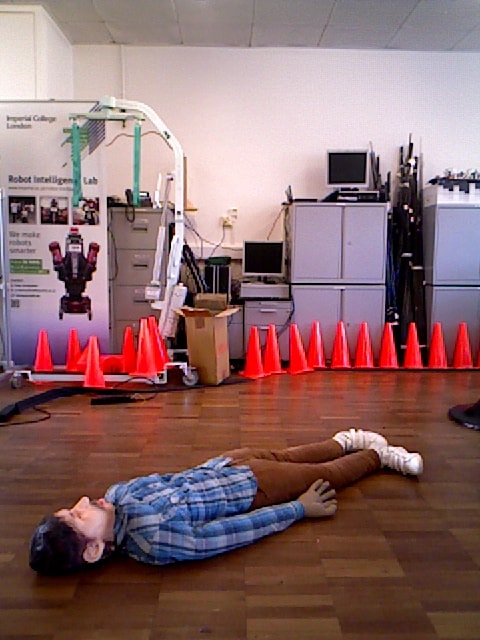}
        \subcaption{}
    \end{subfigure}
    ~~~~
    \begin{subfigure}[c]{0.1\textwidth}
        \includegraphics[width=0.7in]{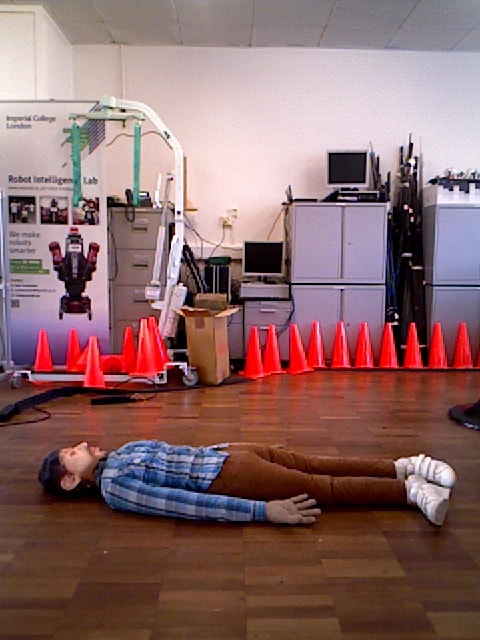} 
        \subcaption{}
    \end{subfigure}
    ~~~~
    \begin{subfigure}[c]{0.1\textwidth}
        \includegraphics[width=0.7in]{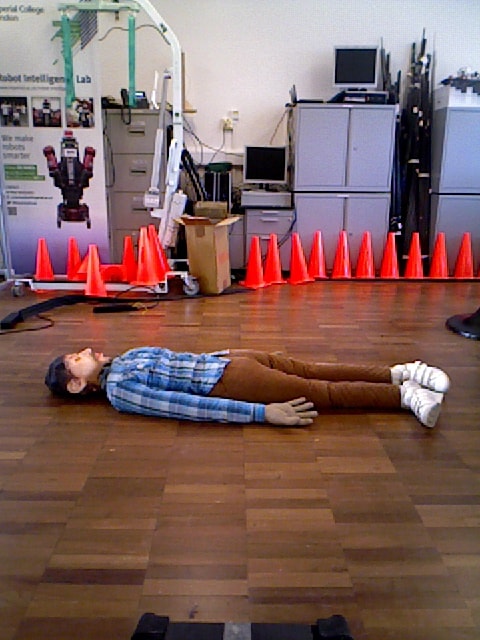}
        \subcaption{}
    \end{subfigure}
    ~~~~
    \begin{subfigure}[c]{0.1\textwidth}
        \includegraphics[width=0.7in]{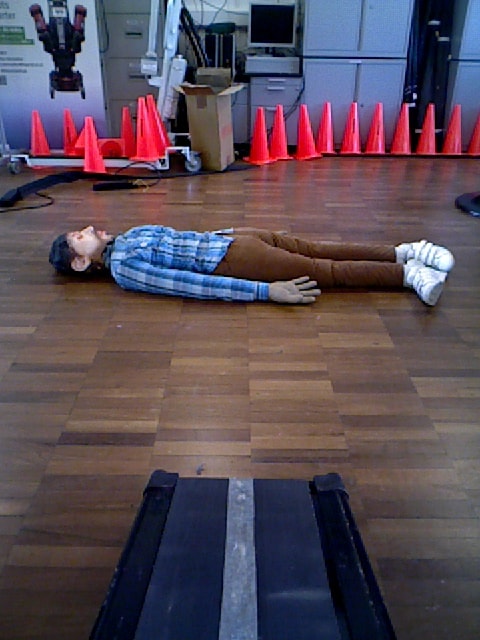}
        \subcaption{}
    \end{subfigure}
    ~~~~
    \begin{subfigure}[c]{0.1\textwidth}
        \includegraphics[width=0.7in]{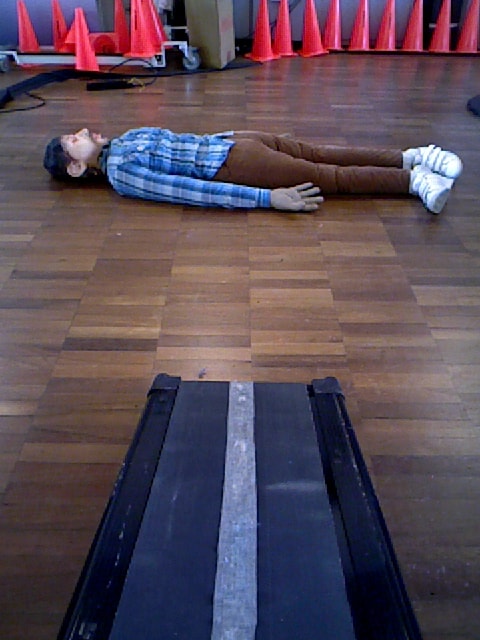}
        \subcaption{}
    \end{subfigure}
    ~~~~
    \begin{subfigure}[c]{0.1\textwidth}
        \includegraphics[width=0.7in]{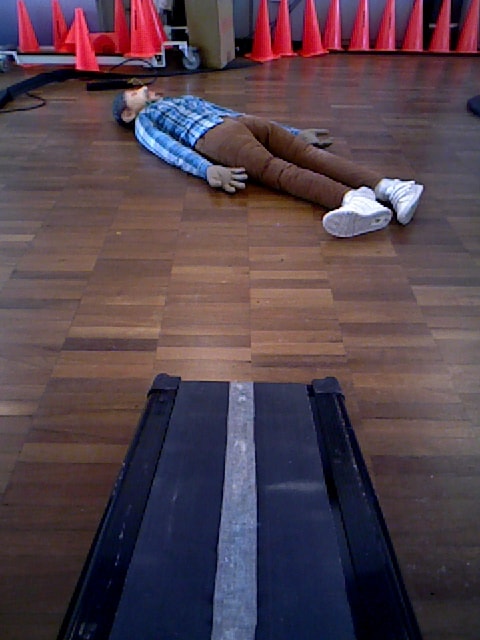} 
        \subcaption{}
    \end{subfigure}
    ~~~~
    \begin{subfigure}[c]{0.1\textwidth}
        \includegraphics[width=0.7in]{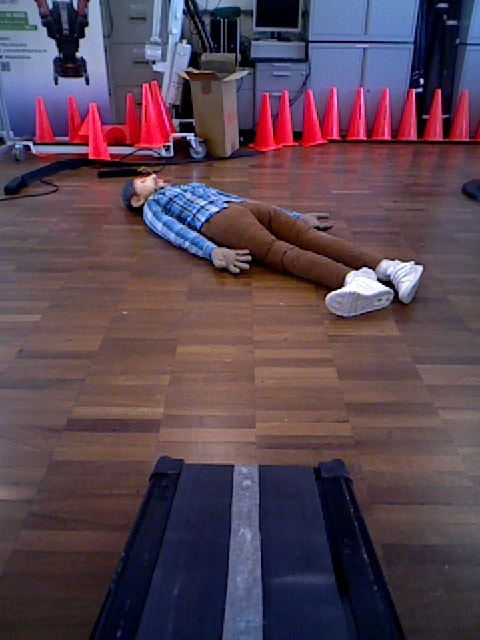}
        \subcaption{}
    \end{subfigure}
    ~~~~
    \begin{subfigure}[c]{0.1\textwidth}
        \includegraphics[width=0.7in]{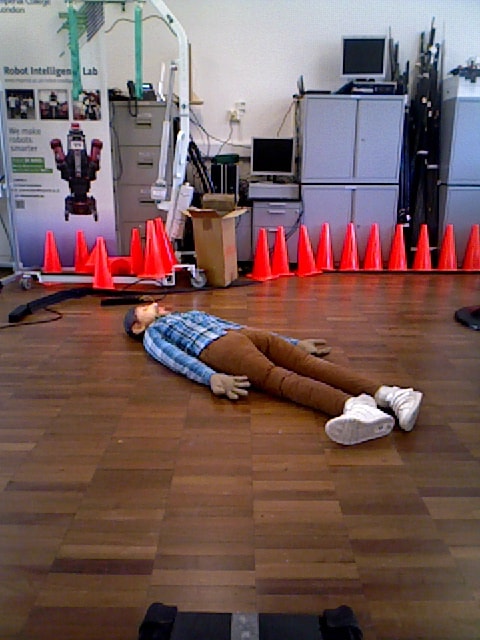}
        \subcaption{}
    \end{subfigure}
    ~~~~
    \begin{subfigure}[c]{0.1\textwidth}
        \includegraphics[width=0.7in]{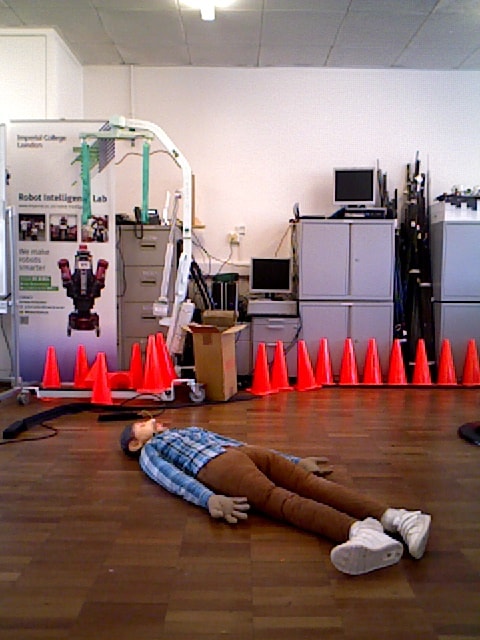}
        \subcaption{}
    \end{subfigure}

    \caption{Sixteen different combinations of the casualty orientations---w.r.t. the onboard RGB-D camera---and the camera angles.}
    \label{fig:sixteen-scenarios}
\end{figure}

\begin{figure}[!h]
    \begin{subfigure}[c]{0.09\textwidth}
        \includegraphics[width=0.8in]{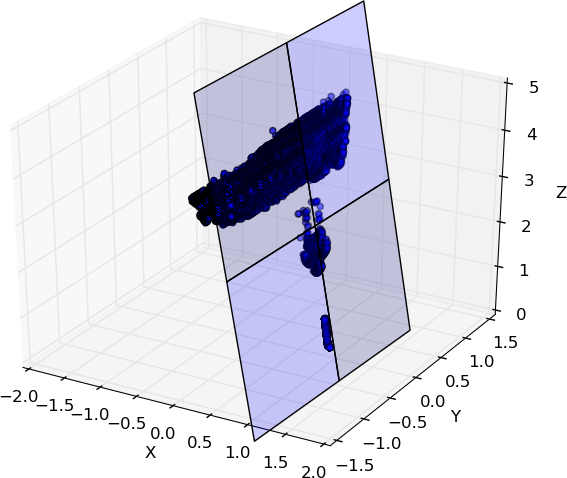} 
        \subcaption{}
    \end{subfigure}
    ~~~~
    \begin{subfigure}[c]{0.09\textwidth}
        \includegraphics[width=0.8in]{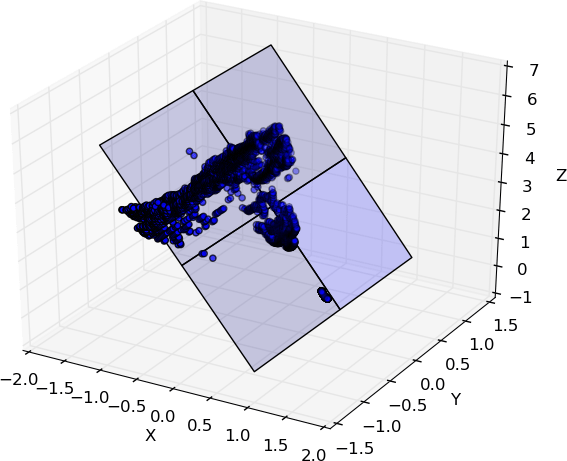}
       \subcaption{}
    \end{subfigure}
    ~~~~
    \begin{subfigure}[c]{0.09\textwidth}
        \includegraphics[width=0.8in]{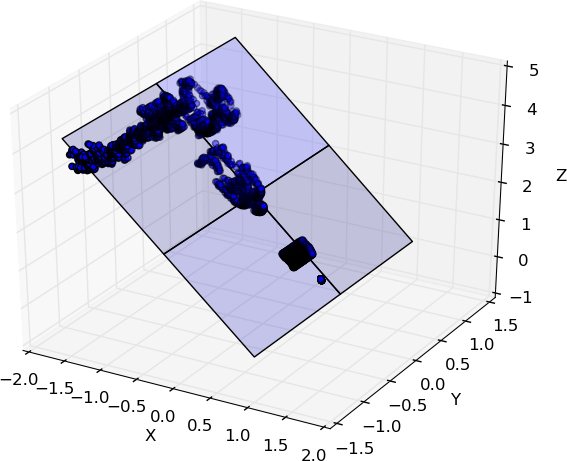}
        \subcaption{}
    \end{subfigure}
    ~~~~
    \begin{subfigure}[c]{0.125\textwidth}
        \includegraphics[width=0.8in]{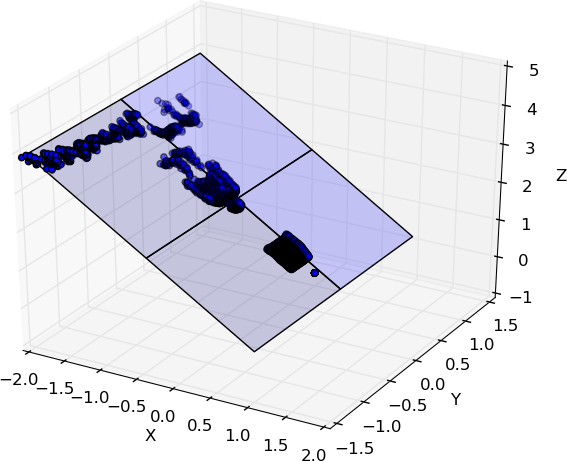}
        \subcaption{}
    \end{subfigure}
    ~~~~
    \begin{subfigure}[c]{0.09\textwidth}
        \includegraphics[width=0.8in]{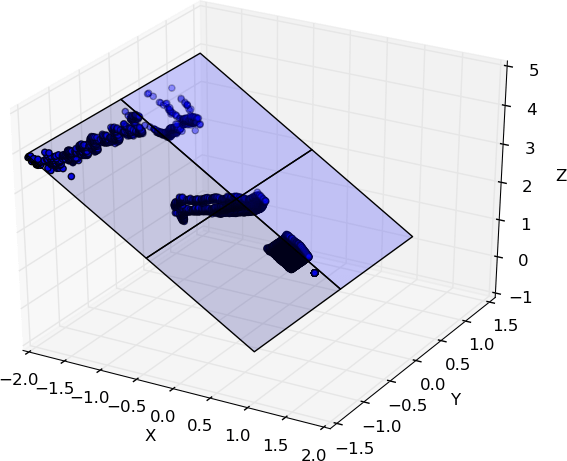} 
        \subcaption{}
    \end{subfigure}
    ~~~~
    \begin{subfigure}[c]{0.09\textwidth}
        \includegraphics[width=0.8in]{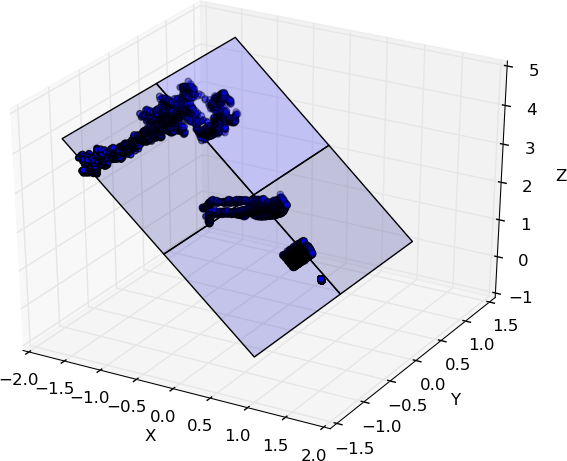}
        \subcaption{}
    \end{subfigure}
    ~~~~
    \begin{subfigure}[c]{0.09\textwidth}
        \includegraphics[width=0.8in]{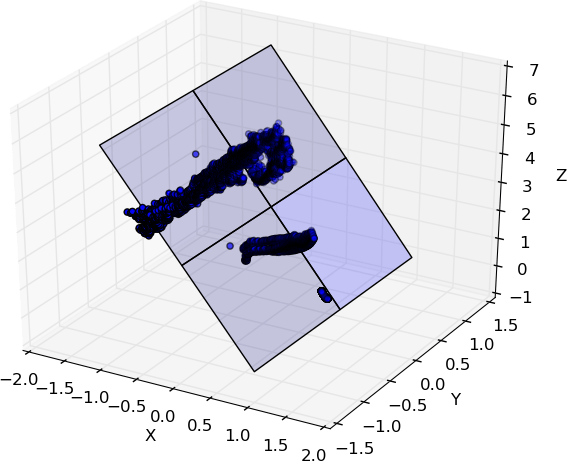}
        \subcaption{}
    \end{subfigure}
    ~~~~
    \begin{subfigure}[c]{0.125\textwidth}
        \includegraphics[width=0.8in]{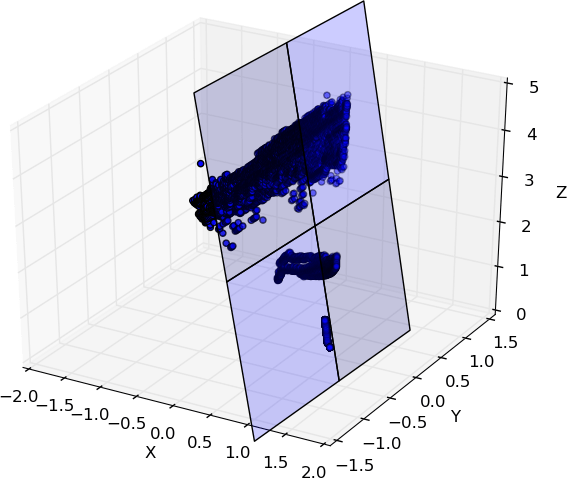}
        \subcaption{}
    \end{subfigure}
    ~~~~
    \begin{subfigure}[c]{0.09\textwidth}
        \includegraphics[width=0.8in]{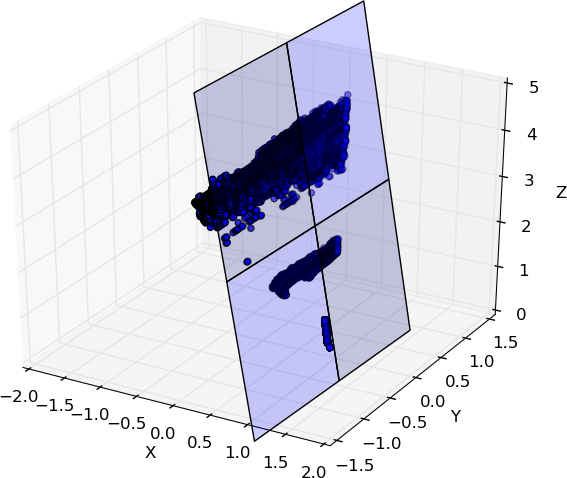} 
        \subcaption{}
    \end{subfigure}
    ~~~~
    \begin{subfigure}[c]{0.09\textwidth}
        \includegraphics[width=0.8in]{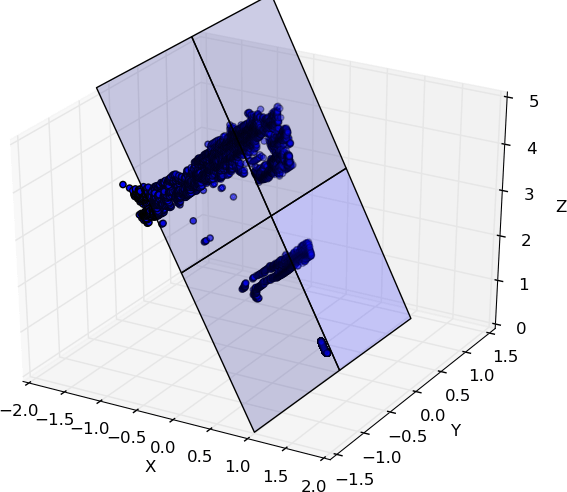}
        \subcaption{}
    \end{subfigure}
    ~~~~
    \begin{subfigure}[c]{0.09\textwidth}
        \includegraphics[width=0.8in]{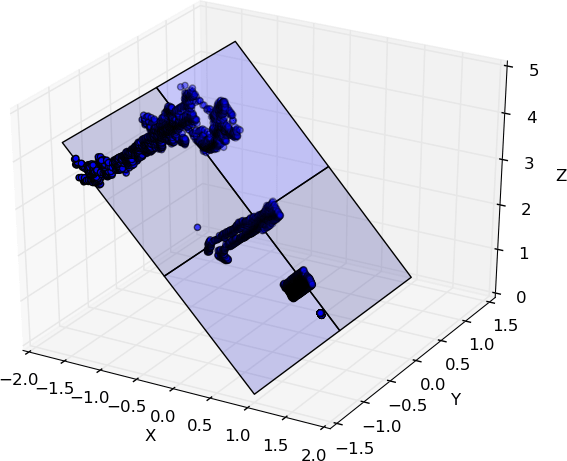}
        \subcaption{}
    \end{subfigure}
    ~~~~
    \begin{subfigure}[c]{0.125\textwidth}
        \includegraphics[width=0.8in]{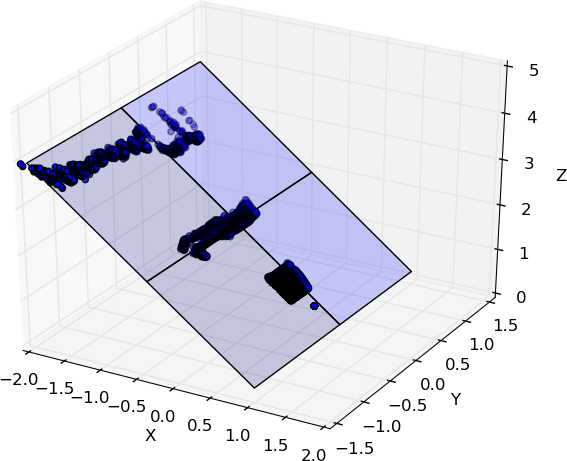}
        \subcaption{}
    \end{subfigure}
    ~~~~
    \begin{subfigure}[c]{0.09\textwidth}
        \includegraphics[width=0.8in]{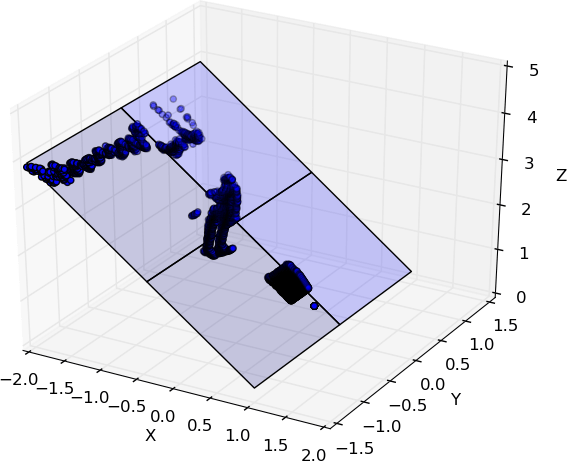} 
        \subcaption{}
    \end{subfigure}
    ~~~~
    \begin{subfigure}[c]{0.09\textwidth}
        \includegraphics[width=0.8in]{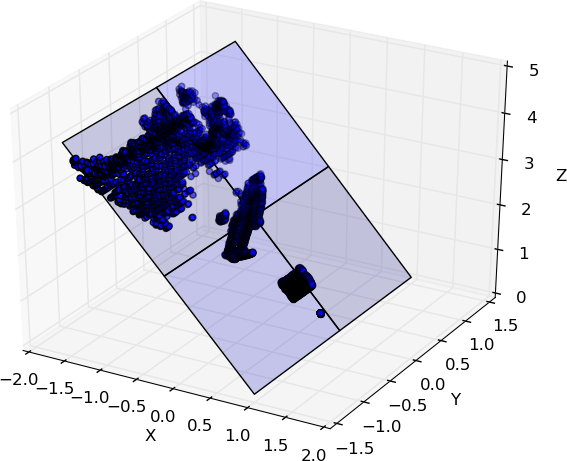}
        \subcaption{}
    \end{subfigure}
    ~~~~
    \begin{subfigure}[c]{0.09\textwidth}
        \includegraphics[width=0.8in]{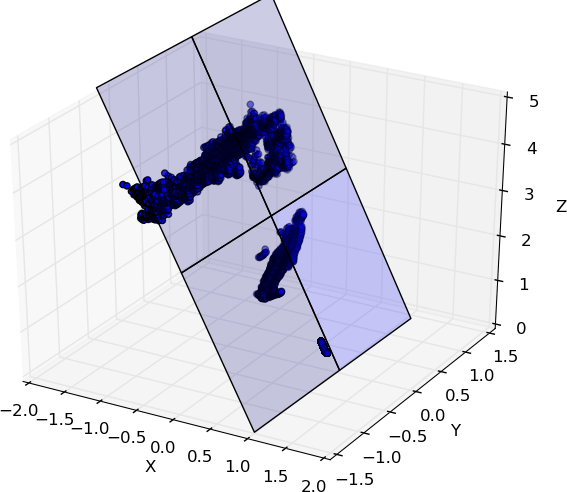}
        \subcaption{}
    \end{subfigure}
    ~~~~
    \begin{subfigure}[c]{0.125\textwidth}
        \includegraphics[width=0.8in]{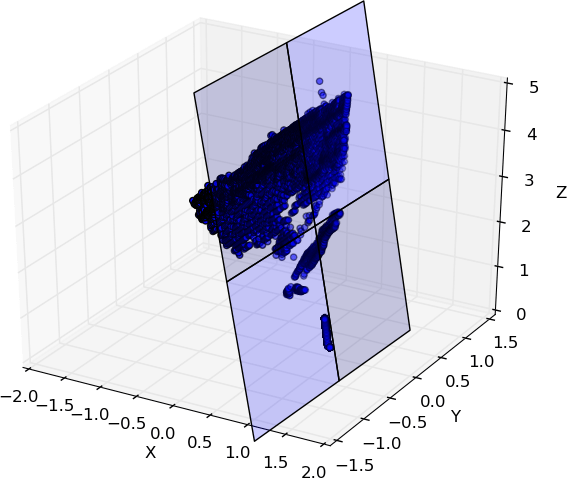}
        \subcaption{}
    \end{subfigure}
    \caption{Projected point clouds on the ground plane with points belonging to the plane removed.}
    \label{fig:projection-results}
\end{figure}

For each camera angle setting, we placed the human dummy in four different random positions and orientations (see Figure~\ref{fig:sixteen-scenarios}). The sixteen RGB images in Figure ~\ref{fig:sixteen-scenarios} illustrate all the human dummy positions and orientations and the camera angle settings used to test the proposed casualty detection method. These scenarios simulate the wide range of variation that the robot sees through the onboard camera. When dealing with these scenarios, standard human detection does not work, but we added these scenarios to retrain the classifier that we used for standard human body detection. The proposed advantages of our approach are that we can use the existing standard human detection method to perform casualty detection in these scenarios.
We obtained the point-cloud data from the on-board RGB-D camera of ResQbot for each tested scenario. Using this point-cloud data, we then performed ground plane detection using the RANSAC algorithm and projected all the points onto the detected plane. Figure~\ref{fig:projection-results} shows the results of the floor detection and point-cloud projection for all sixteen scenarios.

\begin{figure}[h!]
    \begin{subfigure}[c]{0.09\textwidth}
        \includegraphics[width=0.8in]{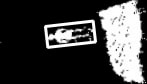} 
        \subcaption{}
    \end{subfigure}
    ~~~~
    \begin{subfigure}[c]{0.09\textwidth}
        \includegraphics[width=0.8in]{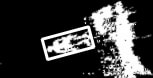}
        \subcaption{}
    \end{subfigure}
    ~~~~
    \begin{subfigure}[c]{0.09\textwidth}
        \includegraphics[width=0.8in]{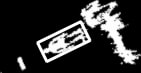}
        \subcaption{}
    \end{subfigure}
    ~~~~
    \begin{subfigure}[c]{0.125\textwidth}
        \includegraphics[width=0.8in]{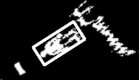}
        \subcaption{}
    \end{subfigure}
    ~~~~
    \begin{subfigure}[c]{0.09\textwidth}
        \includegraphics[width=0.8in]{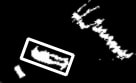} 
        \subcaption{}
    \end{subfigure}
    ~~~~
    \begin{subfigure}[c]{0.09\textwidth}
        \includegraphics[width=0.8in]{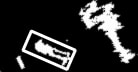}
        \subcaption{}
    \end{subfigure}
    ~~~~
    \begin{subfigure}[c]{0.09\textwidth}
        \includegraphics[width=0.8in]{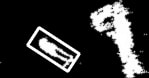}
        \subcaption{}
    \end{subfigure}
    ~~~~
    \begin{subfigure}[c]{0.125\textwidth}
        \includegraphics[width=0.8in]{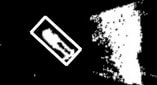}
        \subcaption{}
    \end{subfigure}
    ~~~~
    \begin{subfigure}[c]{0.09\textwidth}
        \includegraphics[width=0.8in]{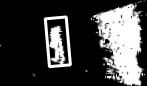} 
        \subcaption{}
    \end{subfigure}
    ~~~~
    \begin{subfigure}[c]{0.09\textwidth}
        \includegraphics[width=0.8in]{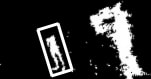}
        \subcaption{}
    \end{subfigure}
    ~~~~
    \begin{subfigure}[c]{0.09\textwidth}
        \includegraphics[width=0.8in]{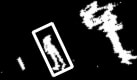}
        \subcaption{}
    \end{subfigure}
    ~~~~
    \begin{subfigure}[c]{0.125\textwidth}
        \includegraphics[width=0.8in]{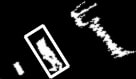}
        \subcaption{}
    \end{subfigure}
    ~~~~
    \begin{subfigure}[c]{0.09\textwidth}
        \includegraphics[width=0.8in]{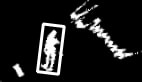} 
        \subcaption{}
    \end{subfigure}
    ~~~~
    \begin{subfigure}[c]{0.09\textwidth}
        \includegraphics[width=0.8in]{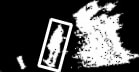}
        \subcaption{}
    \end{subfigure}
    ~~~~
    \begin{subfigure}[c]{0.09\textwidth}
        \includegraphics[width=0.8in]{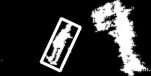}
        \subcaption{}
    \end{subfigure}
    ~~~~
    \begin{subfigure}[c]{0.125\textwidth}
        \includegraphics[width=0.8in]{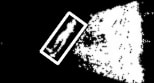}
        \subcaption{}
    \end{subfigure}

    \caption{The results of the generated grid maps and the casualties detected (white rectangle).}
    \label{fig:detection-results}
\end{figure}


The projected point-cloud data of each scenario was then converted into grid maps that were used to perform casualty detection. To detect the casualty, we performed a template-matching method using single samples obtained from the RGB-D camera facing directly toward the human dummy (see Figure~\ref{fig:template}).

Figure~\ref{fig:detection-results} shows the obtained grid maps from the projected point cloud of all sixteen scenarios. It also shows the casualty detection results of the template matching for each scenario. The experimental results demonstrate that the casualty detector can achieve near-perfect performance when detecting the casualty, i.e. the human dummy, in all the scenarios tested. Therefore, the conducted evaluation is proof that the proposed approach has potential for detecting casualties using onboard sensors on a mobile rescue robot.

In addition to the experiments, we also compared our proposed approach with the state-of-the-art of object and human presence detections, including HOG-SVM method [28], pre-trained RetinaNet [29][30] and pre-trained Darknet (i.e. YOLO) [31][32]. We compared the detection performance of these methods to our proposed method in detecting the casualty (i.e. human dummy) in the images that are tested on the experiments, including the template image (i.e. Fig. 4a) and the test images (i.e. Fig. 6a to Fig. 6p). The summary of the comparison results can be seen in the Table~\ref{tab:comparison}.  The results show that while the proposed method succeeded to detect the casualty in all images, the state-of-the-art techniques failed to recognise the casualty in several scenarios. More specifically, these detectors cannot identify the casualty in the situations that the visibility is limited due to the camera angles and the casualty orientations.

\begin{table}[t]
\begin{center}
  \caption{Comparison to the state-of-the-art of object detection methods}
  \label{tab:comparison}
  \begin{tabular}{rcl}
    \toprule
    Method&Template Image&Test Images\\
    \midrule
    HOG + Linear SVM&Detected&$0$ out of $16$\\
    RetinaNet&Detected&$8$ out of $16$ \\
    YOLO&Detected&$12$ out of $16$\\
    Proposed GPPC&Detected&$16$ out of $16$\\
  \bottomrule
\end{tabular}
\end{center}
\end{table}

\section{Conclusion and Future Work}
In this paper, we introduce a novel approach to the casualty detection problem that uses point-cloud data from onboard sensors on a mobile rescue robot. We use a template-matching method to detect objects in images combined with our approach for detecting casualties. We conducted experimental testing for the proposed method to evaluate its ability to detect casualties lying on the ground in several different orientations with respect to the onboard RGB-D camera position. We also tested several different camera angles of the RGB-D camera with respect to the ground plane to evaluate the robustness of our method regarding changes in camera angle. 

The results show that the detector succeed to detect the casualty, i.e. the human dummy, in the all tested scenarios. To deal with these scenarios, the detector only used single grid map template. It demonstrates that the proposed approach is promising as a basis for further extension and development of casualty detection method for a mobile rescue robot.

The ongoing work focuses on developing the detection approach to detect a casualty lying on the ground. In the future, we are planning to extend the evaluation of the proposed casualty detection method by incorporating more extensive scenarios, such as different body configurations and comparisons with the existing state-of-the art methods used for casualty or human-body detection. 

Moreover, we have also developed a mobile rescue robot, ResQbot, that is capable of performing a casualty extraction procedure via teleoperation. We intend to integrate the autonomous casualty detection procedure into ResQbot, so that this robot is able to perform the casualty extraction procedure fully autonomously.





\section*{ACKNOWLEDGMENT}

Roni Permana Saputra would like to thank the Indonesia Endowment Fund for Education -- LPDP, for the financial support of the PhD program.  
The authors also would like to show our gratitude to Nemanja Rakicevic and Ke Wang for sharing their comments and feedback on this work.

\end{document}